# Unifying Geometric Features and Facial Action Units for Improved Performance of Facial Expression Analysis

Mehdi Ghayoumi[1], Arvind K Bansal[1]

[1]*Computer Science Department, Kent State University,*
*{mghayoum,akbansal}@kent.edu*

Abstract: Previous approaches to model and analyze facial expression analysis use three different techniques: facial action units, geometric features and graph based modelling. However, previous approaches have treated these technique separately. There is an interrelationship between these techniques. The facial expression analysis is significantly improved by utilizing these mappings between major geometric features involved in facial expressions and the subset of facial action units whose presence or absence are unique to a facial expression. This paper combines dimension reduction techniques and image classification with search space pruning achieved by this unique subset of facial action units to significantly prune the search space. The performance results on the publicly facial expression database shows an improvement in performance by 70% over time while maintaining the emotion recognition correctness.

## 1 INTRODUCTION

Your Emotion represents an internal state of human mind [28], and affects their interaction with the world. Emotion recognition has become an important research area in: 1) the entertainment industry to assess consumer response; 2) health care industry to interact with patients and elderly persons; and 3) the social robotics for effective human-robot interaction. Online facial emotion recognition or detection of emotion states from video has applications in video games, medicine, and affective computing [26]. It will also be useful in future in auto-industry and smart homes to provide right ambience and interaction with the occupying humans. Emotions are expressed by: (1) behavior [28]; (2) spoken dialogs [22]; (3) verbal actions such as variations in speech and its intensity including silence; (3) non-verbally using gestures; (4) facial expressions [11] and tears; and (5) their combinations. In addition to the analysis of these signals, one has to be able to analyze and understand the preceding events and/or predicted future events, individual expectations, personality, intentions, cultural expectations, and the intensity of an action. There are many studies to classify primary and derived emotions [4, 6, 16, and 28].

During conversation, people scan facial expressions of other persons to get a visual cue to their emotions. In social robotics, it is essential for robots to analyze facial expressions and express a subset of human emotions to have a meaningful human-robot interaction.

There are many schemes for the classification of human emotions. One popular theory for social robotics is due to Ekman [10, 11] that classifies human emotions into six basic categories: surprise, fear, disgust, anger, happiness, and sadness. In addition, there are many composite emotions derived by the combination of these basic emotions. The transitions between emotions that require continuous facial-expression analysis.

Three major techniques have been used to simulate and study human facial expressions: FACS (Facial Action Control System) [5], GF (Geometric Features) and GBMT (Graph Based Modeling Techniques) [7]. FACS simulates facial muscle movement using a combination of facial action units (FAUS or AUs). Different combinations of AUs model different muscle movements and specific facial expressions. FACS has found a major use in realistic visualization of facial-expressions through animation [1, 13]. Facial expression analysis techniques are based upon geometrical feature extraction [8], modeling extracted features as graphs, and analyzing the variations in the graph for deviation.

Previous techniques [9] start afresh every time they analyze the emotion, and the accounting for expectations of emotions is not important. They also do not take into account the fact that a subset of features are unique to the presence or the absence of specific facial-expression. Identification of these subsets of unique features during facial expression analysis can prune the search space.

In this paper, we identify subsets of action units (AUs) that uniquely characterize the presence or the absence of a subset of emotions and map these AUs to the geometric feature-points to prune the search space. The technique extends previous facial expression identification techniques based upon LSH (Locality Sensitive Hashing) [17] that employ LSH for efficient pruning of search space.

The technique has been demonstrated using a publicly available image database [21, 23] that has been used by previous approaches. Results show significant improvement in performance over time (70% improvement in execution time) compared to previous techniques while retaining the accuracy in the similar range. The proposed technique is also suitable for fast emotion recognition in videos and real-time robot-human interaction, as the scheme recognizes emotion transitions.

The major contributions of this paper are:

Applying the subset of action units for pruning the search space for different emotions during interactive communication.

Combining these subsets of action units with geometric modeling to reduce the number of feature points and transformation, thus improving execution efficiency.

The rest of the paper is organized as follows. Section 2 describes the background of FACS, geometric features for emotion, Principal Component Analysis (PCA) and Support Vector Machine (SVM). Section 3 describes the proposed approach. Section 4 refers to the algorithm and the implementation. Section 5 demonstrates the dataset and the results. Section 6 explains the related works and the last section concludes the paper and describes the future works.

## 2. BACKGROUND

### 2.1. FACS - Facial Action Control System

Contractions of a subset of facial muscles generate human facial expressions. A set of 66 AUs (Action Units) [11] have been used to simulate the contractions of facial muscles. An action unit simulates the activities of one or several muscles in the face. Tables 1 and 2 describe a relevant subset of action units needed for the simulation of the facial expressions for the six basic emotions.

**Table 2. Set of action units needed for basic emotions**

| Basic expressions | Involved Action Units |
|---|---|
| Surprise | AU 1, 2, 5,15,16, 20, 26 |
| Fear | AU 1, 2, 4, 5,15,20, 26 |
| Disgust | AU 2, 4, 9, 15, 17 |
| Anger | AU 2, 4, 7, 9,10, 20, 26 |
| Happiness | AU 1, 6,12,14 |
| Sadness | AU 1, 4,15, 23 |

### 2.2. PCA - Principal Component Analysis

PCA is a dimension reduction technique that transforms the data-points to a new coordinate system using orthogonal linear transformation, such that the transformed data-points lie with greatest variance on the first coordinate. Only the dimensions with major variations are chosen for further analysis, reducing the feature space. It is based upon finding out the eigenvectors and eigenvalues [14].

### 2.3. SVM - Support Vector Machine

SVM creates a set of hyper-planes in a high-dimensional space, which can be used for classification, regression, or other tasks. A good separation is achieved by a hyperplane that has the largest distance to the nearest training data point of any class (functional margin). The operation of the SVM algorithm is based on finding the hyperplane that gives the largest minimum distance to the training examples, and it is called *margin*. Support Vectors (SV) are the elements of the training set that would change the position of the dividing hyperplane that are critical elements of the training set and are closest to the hyperplane. In general, the larger margin makes lower generalization error of the classifier [3, 25].

### 2.4. Geometric features in facial expression

Face-features can be modeled as a graph [15, 18]. Face movement is a combination of all facial feature points, but some points have a main role in facial expression. There are three types of nodes (feature-points): *stable, passive* and *active*. Stable feature-points are fixed. Passive feature-points do not have significant muscle movement. Active feature-points are most affected by muscle movements and the change in position of active-points makes the change in facial-expressions.

The number of feature points have been reduced from 62 points to 24 major points without loss of information as these 24 major points are present in all basic emotions. In the modified model, there are 6 points in eyebrows {$bl_1…bl_3$, $br_1…br_3$}, 8 points in eyes {$el_1…el_4$, $er_1…er_4$} and 10 points on mouth {$ml_1…ml_3$, $mm_1…mm_4$, $mr_1…mr_3$}. The subset {$er_1$, $el_1$} represents stable points, the subset {$er_4$, $el_4$, $mr_2$, $ml_2$} represents passive points, and the subset {$br_1$, $br_2$, $br_3$, $bl_1$, $bl_2$, $bl_3$, $er_2$, $er_3$, $el_2$, $el_3$, $mr_1$, $mr_2$, $mr_2$, $ml_1$, $ml_2$, $ml_3$, $mm_1$, $mm_2$, $mm_3$, $mm_4$} represents the active-points. Figure 1 shows the geometric model with the feature points.

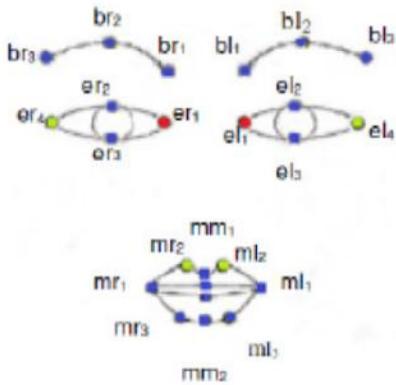

Figure 1. Feature-points in a geometric model of a face

The transformation matrix maps muscles and face movement to a formula based upon the movement of the feature points. Each movement is a combination of translation, rotation, and scaling. This transformation is caused due to head-movements, and the change in coordinates of the fixed inner eye corners $er_1$ with coordinate ($x_r$, $y_r$) and $el_1$ with coordinate ($x_l$, $y_l$). The transformations are given in equations (1) thru (5). The abbreviations *norm, Trans, rot,* and *sc* denote *normalize, transform, rotate* and *scale* respectively.

$$norm(x,y) = sc(x,y) \times rot(x,y) \times trans(x,y) \quad (1)$$

$$trans(x,y) = \begin{bmatrix} -\frac{x_l + x_r}{2} \\ -\frac{y_l + y_r}{2} \end{bmatrix} \begin{bmatrix} x \\ y \end{bmatrix} \quad (2)$$

Where ($x_l$, $y_l$) and ($x_r$, $y_r$) are the coordinates of left and right inner eye corners $el_1$ and $er_1$ respectively:

$$rot(x,y) \begin{bmatrix} \cos(-\theta) & -\sin(-\theta) \\ \sin(-\theta) & \cos(-\theta) \end{bmatrix} \begin{bmatrix} x \\ y \end{bmatrix} \quad (3)$$

Where θ is the angle between the intervals joining the inner eye corners and the horizontal x-axis.

$$sc(x,y) = \frac{1}{2xr} \begin{bmatrix} x \\ y \end{bmatrix} \quad (4)$$

$$sc(x,y) \frac{1}{2xl} \begin{bmatrix} x \\ y \end{bmatrix} \quad (5)$$

Where $x_r$ and $x_l$ are the x-coordinates of right and left eyes respectively.

Table 3 describes the deviations of various facial feature-points that are needed to simulate facial expressions. Various movements of facial feature-points are *left, right, up, down, stretch* and *tighten*.

Table 3. Actions of feature-points in Figure 1

| Feature Points | Deviation |
|---|---|
| Brow points ($br_1$, $br_2$, $br_3$, $bl_1$, $bl_2$, $bl_3$) | up, down |
| Mid points of eyes ($er_2$, $er_3$, $el_2$, $el_3$) | up |
| Outer lip points ($mr_1$, $ml_1$) | stretch, tighten |
| Midpoint of upper lips ($mm_1$, $mm_2$) | up, down |

In order to separate the intensities of feature points for different emotions, the equation for cumulative difference is defined as follows:

$$diff = \sum_{i=1}^{n-1}(E_{i+1} - E_i) - \sum_{i=1}^{n-1}(N_{i+1} - N_i) \quad (6)$$

Where $E_i$ (0 ≤ i ≤ n – 1) represents the feature point of an expressive face-state and $N_i$ (0 ≤ i ≤ n – 1) represents feature point of a neutral face-state respectively. In the equation 6, the outcome *diff > 0* means muscle-elongation and *diff < 0* means muscle-contraction.

### 2.5 New definitions

Facial expressions are modelled using a single action-unit or a composite action-unit made of more than one action-units. For example, the facial expression for "happiness" is characterized by any of the three single AUs: 6, 12, and 14 (see Table 4), while the facial expression for "fear" is defined by a composite action-unit consisting of AUs 4 and 5. Composite action units are modeled as tuples. Thus the facial expression for "fear" is characterized by an AU-tuple (4, 5) (see Table IV).

## 3. APPROACH - UNIFIED METHOD

The integrated method is based upon:

1) Identifying the subset of AUs that are unique to basic emotions as shown in Table IV;
2) The subset of AUs that are absent in basic emotions as shown in Table V;
3) The subset of the AUs that will clearly shows transition from one basic emotion to another emotion as shown in Table VI.

Mapping these subsets to the change in geometric features (see Figure 1) and restricting the runtime check for the changes in the subset of geometric features significantly reduces the execution time of the facial expression analysis. A minimal subset of at least seven AUs are needed to check for the presence of any emotion uniquely. For example, Table 1 shows AU 1, 2, 5, 15, 16, 20 and 26 and for recognition of surprise. However, AU 16 is not needed in other emotions. It means AU 16 is sufficient to recognize the state "surprise". Table IV describes a reduced subset to identify six basic facial expressions. The confidence factor can be improved further by:

1) Checking for additional AUs that characterize facial-expressions as in the case of happiness;
2) Checking for the absence of facial-expressions showing by the presence of AUs (see Table V).

Table 4. Subsets of unique AUs in basic emotions

| State | AUs |
|---|---|
| Surprise | {16} |
| Fear | {(4, 5)} |
| Disgust | {17} |
| Anger | {10} |
| Happiness | {6, 12, 14} |
| Sadness | {23} |

Table 5 lists sets of major AUs for each state. Using these sets, unique subset of action units present/absent in the specific facial expressions can be predicted. For example, about NOT surprise (absence of surprise) is given by the subset {4, 6, and 23}. A reduced subset T = {1, 2, 4, 5, 6, 7, 9} has been used to check for the absence of any uniquely emotion using a simple decision tree as shown in Figure 2.

Table 5. Subsets of AUs absent in basic emotions.

| State | AUs |
|---|---|
| Not surprise (NSur) | {4, 6, 23} |
| Not fear (NF) | {6, 9, 16, 23} |
| Not disgust (ND) | {1, 7} |
| Not anger (NA) | {1, 5, 23} |
| Not happiness (NH) | {2, 4, 5, 9, 10, 16, 17, 20} |
| Not sadness (NSad) | {2, 5, 6, 9, 10, 16, 20} |

To handle the transition of emotions from an existing emotion to another emotion, the tables of the difference between emotions is utilized. Table 6 gives the subsets of actions units that are present or absent when emotion transitions from the state surprise to other basic facial expressions. For example, to derive the transition from the facial expression *fear* to *surprise*, AU 4 should be present, and the subset {7, 9, 10, 17, and 23} should be absent. Similarly, to see transition from the state *surprise* to *happy* any of the three AUs 6 or 12 or 14 are sufficient. The symbol "P" denotes *presence*, and the symbol "A" denotes *absence*.

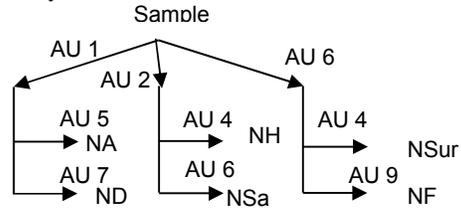

Figure 2. Classification-tree for absence of the emotions.

Table 6. Differences in emotion pairs involving surprise.

| Emotions Pairs | AUs |
|---|---|
| Surprise→ Fear | P: {4};  A: {7, 9, 10, 17, 23} |
| Surprise→ Disgust | P:{4, 9, 17};  A: {9, 10, 23} |
| Surprise→ Anger | P:{4, 7, 9, 10};  A: {17, 23} |
| Surprise→ Happiness | P:{6 / 12 / 14}; A: {4} |
| Surprise→ Sadness | P: {4, 23};  A: {7, 9, 10, 17} |

As shown in the classification tree in Figure 3 by using the minimal subset of {4, 6, 7, 10, 17, and 23} the separation between emotion transitions can be done.

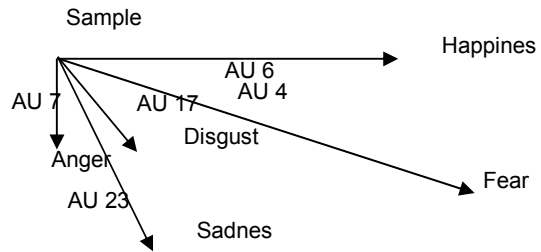

Figure 3. Decision-tree for emotional state after surprise.

### 3.1 Mapping action units to geometric features

Since the image analysis system only sees the changes in the geometric features of the face, the effect of AUs have to be mapped to the observable changes in the geometric features. Table 7 describes a mapping between AUs and the movement of geometric feature points. The mapping shows that many muscle movements map to the same geometric features. For example, AU # 6, 12, and 14 all are involved in stretching $mr_1$ and $ml_1$; AU # 4 and #9 pull down the inner brow points $br_1$ and $bl_1$ down; and AU # 16 and #26 pull $mm_3$ down. . While, multiple emotions may map to the movement of the same feature points, the magnitude of movement is different, and is derived by the SVM training using *diff* equation as explained in section 2.

TABLE 7. MAPPING OF ACTION UNITS TO GEOMETRIC FEATURES

| Action units ↔ Features | | | Action units ↔ Features | | |
|---|---|---|---|---|---|
| AUs | | Features | AUs | | Features |
| AU | Action | Id | Action | AU | Action | Id | Action |
| 1 | up | $br_1$, $bl_1$ | up | 12 | pull | $mr_1$, $ml_1$ | stretch |
| 2 | up | $br_3$, $bl_3$ | up | 14 | dimple | $mr_1$, $ml_1$ | stretch |
| 4 | down | $br_1$, $bl_1$ | down | 16 | down | $mm_3$, $mm_4$ | down |
| 5 | up | $mm_1$, $mm_2$ | up | 17 | up | $mm_3$, $mm_4$ | up |
| 6 | up | $mr_1$, $ml_1$ | stretch | 20 | stretch | $mr_1$, $ml_1$ | stretch |
| 7 | tight | $mr_1$, $ml_1$ | tight | 23 | tight | $mr_1$, $ml_1$ | tight |
| 9 | wrinkle | $br_1$, $bl_1$ | down | 26 | down | $mr_3$, $mm_3$, $ml_3$, | down |

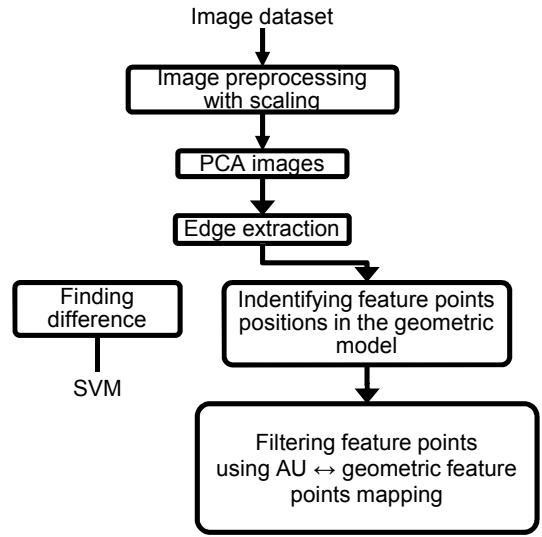

Figure 4. Flow of the algorithm

## 4. IMPLEMENTATION

Facial behaviour of image sequences has been chosen from CK+ database [23]. We compared emotion recognition results on the CK+ database over different dimensions with those produced by SVM. For each emotion category, one-third of the 653 images in the database were selected for training and the remaining images were used for testing. Images of size 490*400 were transformed into 196000*1 dimensional column vectors. Input features sizes are high and a PCA was used to reduce image dimension. All experiments have been implemented using MATLAB. Figure 4 shows the process. For the first step, a part of the dataset will be selected and after that, some image pre-processing such as scaling is applied on images. Then PCA is used to find component images and then canny filter is applied for edge extraction from PCA components. Now the pixels around the geometric feature points should be extracted in each image for finding measures and difference for SVM training. The number of pixels are filtered by using the AU and geometric feature point mapping as explained in Section 3.

The inputs of SVM are some vectors that are related to each image in the database and extracted using AUs and the geometric feature model.

## 5. Experimental results

In the figure 4, the flow of algorithm has presented. At first some processing on the images is done that include some resizing and finding face in each image. Then the PCA is applied on the image that its outputs are some transformed images by eigenvectors. Edge detection of PCA images is the next step. Then, feature points and their position are extracted. In follow some points that are related to the geometric model are selected and finally the differences of extracted pint are calculated for training and testing SVM.

Figure 5 shows a sample of randomly selected emotion-state images after principle component analysis.

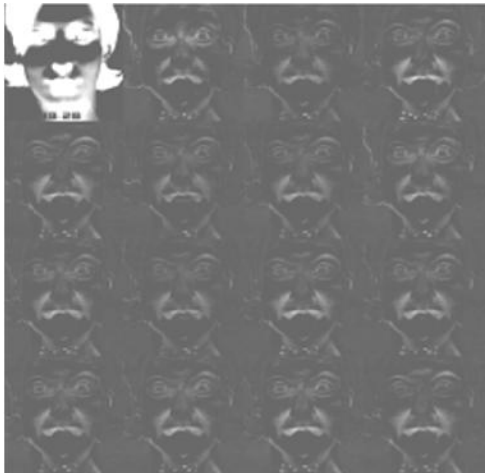

Figure 5. PCA components

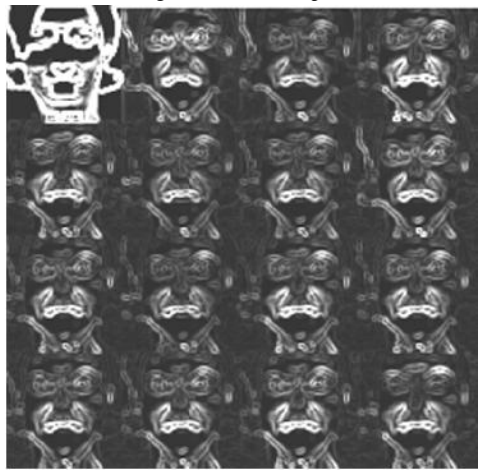

Figure 6 shows the output after the corresponding images after edge extraction.

Table 8 presents the percentage of correctness, number of SVs and margin for emotion recognition by the general approach using just SVM. The abbreviation *NSur* denotes "not surprise", *NA* denotes "not angry", *NSad* denotes "not sad", *NH* denotes "not happy", *ND* denotes "not disgust", and *NF* denotes "not fear".

Table 8. Emotion recognition using just SVM

| Emotions | SVnumber | Margin | Correctness |
|---|---|---|---|
| NSur | 8.1 | 1.95 | 75 % |
| NA | 7.3 | 1.83 | 73% |
| NSa | 8.3 | 1.88 | 76% |
| NH | 6.5 | 1.95 | 84% |
| ND | 8.1 | 1.79 | 79% |
| NF | 7.3 | 1.83 | 74% |

Table 9 shows the correctness using the proposed unified approach. It is clear in Table 9 that correctness is significantly better for *NSur and NA*, and it is comparable for *NSad, NH, ND* and *NF*.

Table 9. Emotion recognition using proposed method.

| Emotions | SV number | Margin | Correctness |
|---|---|---|---|
| NSur | 7.1 | 1.67 | 83 |
| NA | 7.9 | 1.60 | 81 |
| NSa | 7.5 | 1.88 | 74 |
| NH | 7.3 | 1.56 | 82 |
| ND | 7.6 | 1.70 | 75 |
| NF | 8.8 | 1.83 | 78 |

Based on the results shown in the Tables 8 and 9, the average SV number is 7.6 in previous method and 7.7 in the proposed method. The average margin is 1.83 in previous method and 1.71 in the method described in this paper. Clearly, the new method is more time-efficient. Table 10 compares the execution efficiency of different approaches. It is clear that with this strategy processing time improves by 70% due to the reduction of the number of AUs and the corresponding feature-points in a face.

Table 10. Execution efficiency of the proposed method.

| Emotions | Execution time (old method) | Execution Time (new method) |
|---|---|---|
| NSur | 7.5 | 1.3 |
| NA | 6.7 | 1.3 |
| NSa | 7.3 | 1.3 |
| NH | 6.7 | 1.3 |
| ND | 7.4 | 1.3 |
| NF | 7.5 | 1.8 |

## 6. Related works

FACS system has been used for emotion generation by many researchers using AU based simulation [1, 13]. Many researchers use a geometric model [18, 29, and 31] and try to improve geometric models. Some articles present a framework for recognition of facial action unit (AU) combinations by viewing the classification as a special representation problem [29, 31], and others present heuristic methods for achieving better performance [23, 27, 30, 32]. Here a modified geometric model has been used that reduces the facial points and uses another metric for finding distances. In some researches, a modified coding system has been presented [19], and some others' work researches integrate coding system, improved strategies and methods [20]. There are many situations that need to recognize an emotion that does not exist in the coding tables. In the real world and implementing algorithms we compare or combine some other coding or emotion state to find that special state. This article presents a coding system based on the basic emotions that can apply directly in such these situations.  Many researchers have developed the version of a computer vision system that are sensitive to subtle changes in the face [21, 24]. In this paper, we use some statistical method to reduce the dimension of training space. Our proposed scheme significantly improves the performance while retaining accuracy and is suitable for real-time analysis of facial expressions and for real-time human-robot interaction.

## 7. Conclusion and future work

In this paper a unified method for facial expression detection has been presented.  The technique maps the AUs for specific emotions to geometric feature point movements, and uses the characterizing feature points based upon AUs mapping to prune the number of pixels being processed, improving the execution time.  Actually, here the correctness improves or are same in three of the emotions, and are within the range of traditional techniques for the remaining three emotions.  The major gain is the 70% performance improvement over time due to pruning of the number of pixels being processed.  The improved performance makes it suitable for real-time robot-human interaction. This performance in the database with the large number of image or images with the high dimension makes it more applicable.  The scheme can be further improved by analyzing the duration of various emotions when the emotion does not change, and needs minimal analysis.  We are extending the current scheme to incorporate the duration of various emotions.  We are also extending the proposed scheme to handle secondary emotions.